\documentclass{article}

\usepackage[preprint]{neurips_2026}

\usepackage[utf8]{inputenc} 
\usepackage[T1]{fontenc}    
\usepackage{hyperref}       
\usepackage{url}            
\usepackage{booktabs}       
\usepackage{amsfonts}       
\usepackage{nicefrac}       
\usepackage{microtype}      
\usepackage{xcolor}         

\usepackage{microtype} 
\usepackage{booktabs}  
\usepackage{url}  
\usepackage{csquotes}
\usepackage{dirtree}
\usepackage{pifont}
\usepackage{subcaption}
\usepackage{standalone}
\usepackage{tikz}
\usepackage{pgfplots}
\usepackage{float}
\usepackage{amsmath}
\usepackage{amsthm}
\usepackage{threeparttable}
\usepackage{multirow}
\usepackage{graphicx}
\usepackage{amsmath,amsfonts}
\usepackage{bm}
\usepackage{listings}
\usepackage{xcolor}
\usepackage{wrapfig}

\usepackage{adjustbox}
\usepackage{threeparttable}
\usepackage{makecell}

\usepackage{natbib}
\newcommand{\approach}{\texttt{LLMSYS-HPOBench}}

\newcommand{\repolink}[0]{\texttt{\textcolor{blue}{\url{https://github.com/ideas-labo/llmsys-hpobench}}}}

\definecolor{codebg}{RGB}{245,247,250}
\definecolor{codekw}{RGB}{0,70,140}
\definecolor{codecm}{RGB}{90,110,90}
\definecolor{codestr}{RGB}{140,60,20}

\lstdefinestyle{pythonapi}{
    language=Python,
    basicstyle=\ttfamily\footnotesize,
    keywordstyle=\color{codekw}\bfseries,
    commentstyle=\color{codecm},
    stringstyle=\color{codestr},
    backgroundcolor=\color{codebg},
    showstringspaces=false,
    frame=single,
    framerule=0.3pt,
    breaklines=true,
    columns=fullflexible,
    xleftmargin=6pt,
    xrightmargin=6pt
}

\lstdefinestyle{pythonapicompact}{
    style=pythonapi,
    basicstyle=\ttfamily\scriptsize\linespread{0.95}\selectfont,
    aboveskip=4pt,
    belowskip=4pt,
    lineskip=-0.2pt,
    xleftmargin=4pt,
    xrightmargin=4pt
}

\usepackage{filecontents}

\title{{LLMSYS-HPOBench}: Hyperparameter Optimization Benchmark Suite for Real-World LLM Systems}

\newcommand*\samethanks[1][\value{footnote}]{\footnotemark[#1]}

\author{%
  Siyu Wu\thanks{Siyu Wu and Yulong Ye are co-first authors.} \\
  UESTC\\
  China\\
  \texttt{wusy5811@gmail.com} \\
  \And
  Yulong Ye\samethanks[1] \\
  IDEAS Lab \\
  University of Birmingham, UK \\
  \texttt{yxy382@student.bham.ac.uk} \\
  \And
  Zezhen Xiang \\
  UESTC\\
  China\\
   \texttt{zz.xiang.work@gmail.com} \\
  \And
  Pengzhou Chen \\
   UESTC\\
  China\\
  \texttt{cc15523016531@gmail.com} \\
  \And
  Gangda Xiong \\
  UESTC\\
  China\\
   \texttt{gangdaxiong0207@gmail.com} \\
   \And
  Tao Chen\thanks{Tao Chen is the corresponding author.} \\
  IDEAS Lab \\
  University of Birmingham, UK \\
  \texttt{t.chen@bham.ac.uk} \\
}

\begin{document}

\maketitle

\begin{abstract}

Large Language Model (LLM) systems have been the frontier of AI in many application domains, leading to new challenges and opportunities for hyperparameter optimization (HPO) for the AutoML community. However, this type of system exhibits an unprecedented compound space of hyperparameter configuration from both the AI and non-AI components; rich and nonlinear implications from the fidelity factors; and diverse costs of measuring hyperparameter configurations, none of which have been fully captured in existing benchmarks. This paper presents the first (live) benchmark suite and datasets for HPO of real-world LLM systems, dubbed \approach, covering data related to the inference objective values of hyperparameter configurations profiled from running the LLM systems. Currently, \approach~contains 364,450 hyperparameter configurations with a dimensionality of 12--23, 3--5 dimensions of fidelity factor leading to 932 settings, 3--9 inference objective metrics, and 2--10 cost metrics, together with generated logs from measuring the LLM systems. What we seek to advocate is not only a revalidation of the existing HPO algorithms over the frontier LLM systems, but also to provide an evolving platform for the AutoML community to explore new directions of research in this regard. The benchmark suite has been made available at: \repolink.

\end{abstract}

\section{Introduction}\label{section:introduction}




The frontier of AI has evolved from machine/deep learning models into large language models (LLMs) that are often combined with planning, memory, retrieval, and tool usage~\cite{DBLP:journals/corr/abs-2502-18635}, dubbed the system of LLM, or simply the LLM systems. Their real-world effectiveness and efficiency, however, are governed not only by core LLM capabilities, but also by the vast other system components that interact with the LLM, e.g., tools, database, and hardware containers. This leads to a multifaceted hyperparameter space, including the AI parts (e.g., LLM temperature and prompting choices) and non-AI parts (e.g., retrieval and tooling strategies)\footnote{We treat the AI hyperparameters as those that are directly attached to the underlying LLM and can influence its behaviors; otherwise they are non-AI hyperparameters.}~\cite{DBLP:conf/mlsys/AgrawalKMPKGRT24}. It has been shown that different hyperparameter configurations of LLM systems can lead to substantially different inference behaviors in terms of task quality, latency, robustness, and monetary cost~\cite{DBLP:journals/corr/abs-2502-18635}. For example, in retrieval-augmented generation (RAG), increasing retrieval top-$k$ items and enabling a stronger reranker can improve response quality, but often cause extra query time and token cost due to longer context and extra LLM calls~\cite{DBLP:journals/corr/abs-2502-18635}.



The inherent black-box nature of LLM systems and their hyperparameter characteristics lead to a great opportunity for the AutoML community to conduct hyperparameter optimization (HPO), as shown in recent work ~\cite{Brookes2025EvolvingEA}. However, despite the increasing importance of optimizing LLM systems' hyperparameters, we have yet to witness considerable research progress/innovation, primarily because the measurement of hyperparameter configurations over the LLM systems is extremely expensive and effort-heavy: measuring one hyperparameter configuration may require tens of minutes to hours and a non-trivial API/compute budget; scaling to thousands of configurations can consume hundreds of GPU-hours~\cite{ICLR2025_a38cfba4,Kim2025TheCO}. This is in addition to the complexity of deploying real LLM systems.

The above is a corollary of lacking dedicated benchmark suite/datasets on configuration and HPO for LLM systems. Indeed, existing benchmarks/datasets for AutoML~\cite{DBLP:conf/icse/LiX0WT20,eggensperger2021hpobench,DBLP:conf/automl/PfistererSMBB22,DBLP:conf/nips/BansalSJZH22} and for computing systems~\cite{DBLP:journals/corr/abs-2511-16882} cannot fulfill the needs of HPO for LLM systems due to several reasons: 

\begin{itemize} 
    \item They primarily focus on either classic machine/deep learning models or computing systems, which fail to capture the unique structure, AI/non-AI hyperparameter interaction, and fidelity implication for LLM systems. For example, LLM temperature and backend scheduling policy can significantly, sparsely, and jointly affect both quality and latency in a LLM system, while many classic HPO benchmarks assume a model-centred pipeline~\cite{DBLP:conf/icml/YingKCR0H19,DBLP:conf/nips/Pineda-ArangoJW21} or are systems only~\cite{DBLP:journals/corr/abs-2511-16882}.
    \item They focus on simple fidelity settings, while the tasks that LLM systems deal with are often much more complex due to the presence of natural language-based tasks. For example, the fidelity may vary with query difficulty, prompt length, and domain shift, which can non-monotonically influence the cost and objectives. In contrast, classic HPO benchmarks~\cite{DBLP:conf/icml/YingKCR0H19, eggensperger2021hpobench} often assume a monotonic relationship and are limited to dataset size or training epochs.
    \item The diverse costs of measuring every hyperparameter configuration, which can differ for LLM systems, are often limited in existing benchmarks. For example, the cost of LLM systems can include build time and the query tokens required to complete a set of tasks.
\end{itemize}






To bridge this gap, this paper presents \approach, the first live benchmark suite with measured hyperparameter configurations and their inference results over real-world LLM systems, together with other relevant data, e.g., hardware and logs. At the time of writing this paper, \approach~contains data collected by running and profiling seven LLM systems that are of various categories: LLM agent, retrieval-augmented generation (RAG) with LLM, and LLM inference engine, leading to 364,450 hyperparameter configurations with a dimensionality of 12--23, 3--5 fidelity factors of 932 settings, 2--9 objective metrics and 2--10 cost metrics, together with hardware metrics and generated logs from measuring the LLM systems. What we advocate to provide the AutoML community is not merely an efficient suite to compare the better or worse among the HPO algorithms for LLM systems, but also paving the way toward new research opportunities and directions. It is worth noting that preparing and collecting \approach~requires great efforts and resources: by May 2026, the data available are the results of running on 10 servers with $\approx2{,}277{,}600$ CPU core-hours and $\approx94{,}900$ GPU-hours. Yet, \approach~is a live benchmark suite where new data/LLM systems are continually incorporated. \approach~can be accessed at: \repolink.


In the remainder of the paper, we start by presenting the preliminaries and discuss how \approach~differs from the related work in Section~\ref{sec:preliminaries}. We specify the creation procedure of \approach, together with its dataset formats and API access in Section~\ref{section:aipernet}. This then follows a case study of using \approach~to run the well-known HPO algorithms, leading to some new insights in Section~\ref{section:case_study}. Finally, in Section~\ref{section:discussion}, we discuss how \approach~can assist in shaping the future research opportunities for AutoML community before concluding the paper.



    
\section{Preliminaries}
\label{sec:preliminaries}

\subsection{Black-box Hyperparameter Optimization}
\label{subsec:config_tuning}

Black-box hyperparameter optimization (HPO) aims to identify a hyperparameter configuration that optimizes target performance objective(s) when the objective function can only be accessed through evaluation of the model, without gradient information or exploitable structure assumptions. Formally, given a hyperparameter space $\bm{\mathcal{X}}$, the problem is defined as:


\begin{equation}
 \bm{x}^* = \arg\min_{\bm{x \in \mathcal{X}}} f(\bm{x}) \text{ or }  \bm{x}^* = \arg\max_{\bm{x \in \mathcal{X}}} f(\bm{x}),
\end{equation}
where $\bm{x} = (x_1, x_2, \dots, x_n)$ denotes a concrete hyperparameter configuration, and $f(\bm{x})$ represents the performance objective of interest, e.g., error or runtime. In general, $\bm{\mathcal{X}}$ may comprise continuous, integer, ordinal, and categorical hyperparameters. This is a well-formulated problem that has been tackled by the AutoML community~\cite{eggensperger2021hpobench,SMAC,Cowen-Rivers2022-HEBO} and the configurable software/system community~\cite{DBLP:journals/tse/Nair0MSA20, 10.1145/3127479.3128605}. Such a formulation naturally applies to LLM systems, whose behavior is governed by a rich set of tunable hyperparameters, such as model selection, prompting strategies, and memory mechanisms. 

A practical extension of black-box HPO is multi-fidelity hyperparameter optimization, which improves search efficiency by allowing queries at cheaper, lower-fidelity settings. Instead of evaluating a configuration $\bm{x}$ only under the target setting, the algorithm may query $f(\bm{x, z})$, where $\bm{z}=(z_1, z_2, \dots, z_m)$ denotes a fidelity setting instantiated from multiple fidelity factors ($z_i$), and each $z_i$ controls one aspect of evaluation fidelity. Typical examples include dataset subsets~\cite{DBLP:conf/aistats/KleinFBHH17,DBLP:conf/aaai/Hu0TYCD19}, reduced training epochs~\cite{DBLP:journals/jmlr/LiJDRT17,DBLP:conf/ijcai/DomhanSH15}, and feature subsets~\cite{DBLP:journals/jmlr/LiJDRT17}. 
Similar cases can also be found under LLM systems, i.e., shorter prompt length, reduced dataset subset, or fewer document chunks. These lower-fidelity settings may reduce evaluation cost, albeit at the expense of approximation accuracy.

\subsection{Related Work}
\label{subsec:related_work}
In the following, we review two lines of benchmark efforts most relevant to this work. 


\paragraph{\textbf{HPO benchmarks for machine/deep learning models.}} In AutoML, benchmark development has progressed from queryable NAS resources to broader HPO. Early efforts such as \texttt{NAS-Bench-101}~\cite{DBLP:conf/icml/YingKCR0H19} and \texttt{NAS-Bench-201}~\cite{DBLP:conf/iclr/Dong020} precomputed architecture evaluations to enable fast and reproducible NAS comparison across fixed search spaces and datasets. Subsequent benchmark datasets moved beyond architecture search: \texttt{HPO-B}~\cite{DBLP:conf/nips/Pineda-ArangoJW21} organized a large-scale OpenML-based benchmark for black-box HPO, while \texttt{HPOBench}~\cite{eggensperger2021hpobench} provided a unified suite with particular emphasis on multi-fidelity HPO, including raw, tabular, and surrogate variants. This line was further enriched by \texttt{YAHPO Gym}~\cite{DBLP:conf/automl/PfistererSMBB22}, which offers a large surrogate-based collection for multi-fidelity and multi-objective HPO, and \texttt{JAHS-Bench-201}~\cite{DBLP:conf/nips/BansalSJZH22}, which targets joint architecture and hyperparameter search and explicitly supports cost-aware and multi-fidelity studies. In parallel, \texttt{NAS-Bench-Suite}~\cite{DBLP:conf/iclr/MehtaWZKZMSYH22} improved interoperability by exposing diverse NAS benchmarks through a common interface, and \texttt{AMLB}~\cite{DBLP:journals/jmlr/GijsbersBCLPTBV24} provides a standardized benchmark for end-to-end evaluation of AutoML frameworks under controlled tasks and budgets.

\paragraph{\textbf{HPO benchmarks for software systems.}}
Existing benchmark efforts for software system HPO are more heterogeneous and have been widely used~\cite{DBLP:journals/tosem/ChenLBY18,DBLP:journals/tse/ChenCL24,DBLP:conf/sigsoft/Gong023,DBLP:journals/pacmse/0001L24,DBLP:conf/sigsoft/0001L21,DBLP:conf/icse/Ye0L25}. Some benchmarks focus on a specific class of systems. For example, \texttt{CATBench} targets compiler autotuning and captures the complexities of compiler optimization~\cite{torring2025catbench}. Others collect data from a broader range of configurable systems, spanning databases, web services, cloud systems, and video encoders; among them, some consider only a single workload/environment per system~\cite{DBLP:conf/kbse/XiongC25,DBLP:conf/icse/WeberKSAS23,DBLP:conf/icse/HaZ19,DBLP:journals/tse/Nair0MSA20,DBLP:conf/icse/XiangChen26,DBLP:journals/tse/GongCB25}, whereas others explicitly incorporate multiple workloads/environments~\cite{muhlbauer2023analysing,DBLP:journals/jss/LesoilABJ23,DBLP:journals/tse/KrishnaNJM21,DBLP:journals/jss/CaoBWZLZ23,DBLP:journals/pacmse/Gong024}. More broadly, \texttt{MOOT}~\cite{DBLP:journals/corr/abs-2511-16882} curates a large repository of real multi-objective optimization tasks spanning various domains, such as software configuration, cloud tuning, project health, and process modeling, among others.

\paragraph{\textbf{How does \approach~differ.}} Overall, unlike \approach, the prior benchmarks are not designed for the HPO of LLM systems. Notably, the proposed \approach~is distinguished by the following unique characteristics (see Appendix~\ref{app:benchmark_comparison} for more details):

\begin{itemize}
 \item \textbf{Real-world LLM systems (C1):} \approach~contains the real-world LLM systems hosted at the GitHub---a type of artefacts that synergizes both the AI and systems (non-AI).
 
    \item \textbf{Rich and complex fidelity (C2):} For each subject system, \approach\ defines a multi-factor fidelity space with up to 5 fidelity factors related to the tasks, whose combinations induce nearly a thousand executable fidelity settings with non-monotonic influences. This substantially expands the fidelity design space beyond prior multi-fidelity HPO benchmarks, which typically use only one or two fidelity factors under the assumption of monotonic relationships (e.g., using more data incurs higher cost while yielding higher fidelity)~\cite{DBLP:conf/icml/FalknerKH18,DBLP:conf/icml/YingKCR0H19,DBLP:conf/iclr/Dong020,eggensperger2021hpobench}, and the single fidelity system benchmarks~\cite{DBLP:journals/corr/abs-2511-16882}.
    \item \textbf{Heterogeneous configuration space (C3):} Each system exposes a mixed and heterogeneous hyperparameter space, spanning AI components (e.g., model and inference settings) and non-AI components (e.g., scheduling and storage/runtime controls). In the current state of the benchmark, subject systems contain a dimension of 5--8 and 4--17 for AI and non-AI hyperparameters, respectively, leading to more than 364K hyperparameter configurations in total, considering the fidelity settings.
    \item \textbf{Many-objective metrics (C4):} \approach\ records up to nine objective metrics for each measured configuration, covering complementary aspects of LLM system behavior, such as quality-related metrics (e.g., F1-score and success ratio) and performance metrics (e.g., throughput and time-to-first-token). This enables richer evaluation and supports trade-off analysis in HPO for LLM systems, and is competitive with the existing benchmarks with only two datasets considering up to 12 objectives.
    \item \textbf{Multi-dimensional measurement cost (C5):} \approach~records explicit measurement cost up to 10 dimensions, including fine-grained benchmark duration, token usage, and the efforts required to prepare/initialize the LLM system. This goes considerably beyond the other benchmarks, which have merely 1--5 dimensions.
    \item \textbf{Hardware utilization metrics (C6):} \approach\ records hardware statistics for each evaluation/measurement, such as CPU/GPU and memory usage, supporting the analysis of execution efficiency and system behaviors. This is often missing in existing benchmarks.
    \item \textbf{Execution logs (C7):} \approach\ records execution log for each evaluation, supporting inspection and debugging analysis. No current benchmarks contain such information.
\end{itemize}

\subsection{Why AutoML needs HPO for LLM Systems?}


Firstly, LLM systems have become a major form of modern AI applications~\cite{DBLP:conf/icse/Du0WWL0FS0L24,DBLP:conf/aaai/JinLZY26,DBLP:conf/nips/WangJYWOSK024,DBLP:conf/nips/KimLKW24}. While earlier AI research was primarily focused on conventional machine/deep learning models, many recent research directions of AI and real-world applications are centered around LLM systems. This makes HPO for LLM systems an important and timely problem for AutoML.

Secondly, unlike classic machine/deep learning models, the behaviors of LLM systems can be jointly shaped by many design choices, including interaction between AI and non-AI hyperparameters; exhibit complex implications from the fidelity settings; and involve diverse costs in measurement. All those create new challenges and opportunities for building more specialized AutoML algorithms.

Thirdly, AutoML's HPO algorithms are naturally well-positioned to address this challenge. Although further specializations are necessary, their current advances in black-box optimization, surrogate modeling, and transfer learning provide a strong methodological foundation for HPO of LLM systems. 

What remains missing, however, is a benchmark suite that renders this problem systematically measurable, comparable, and reproducible. This is precisely the goal of \approach.
\section{\approach: A Benchmark Suite of Datasets}\label{section:aipernet}



\approach\ is a benchmark suite of datasets for HPO research on LLM systems. By organizing different aspects of relevant information in a unified schema, \approach\ turns expensive end-to-end system evaluations into an efficient, reusable, and reproducible empirical resource for the broader AutoML research on HPO of LLM systems.



\subsection{Benchmark Dataset Construction Procedure}\label{subsection:dataset_procedure}
Figure~\ref{fig:workflow} illustrates the construction pipeline of \approach~for LLM systems HPO. We start by selecting open-sourced real-world LLM systems from GitHub based on timeliness, stars, and completeness of the documentation. For each LLM system, we perform the following steps:



\begin{figure*}
    \centering
    \includegraphics[width=\linewidth]{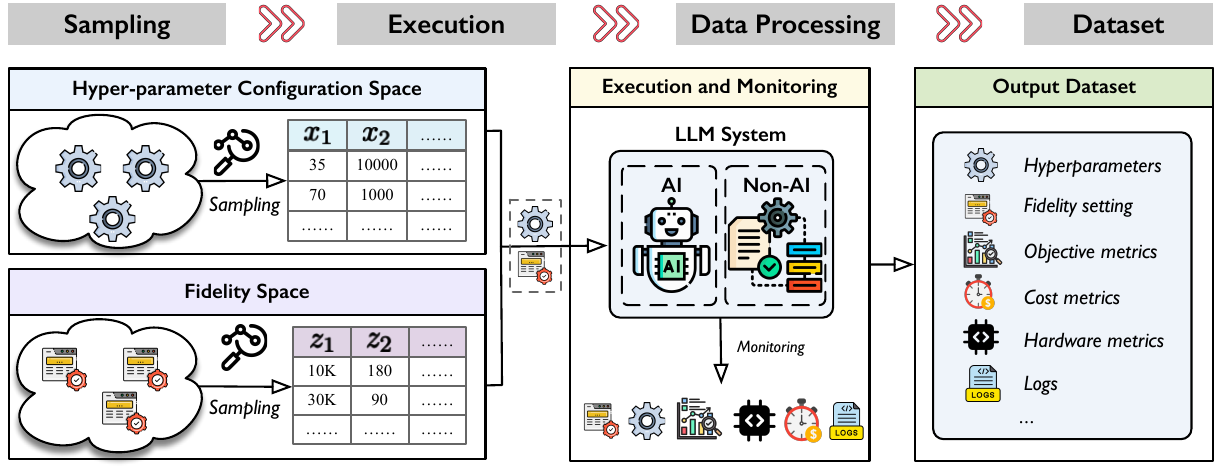}
    \caption{Workflow for constructing \approach.}
    \label{fig:workflow}
\end{figure*}

\begin{table*}[t!]
	\centering
	\caption{The LLM systems profiled so far.}
	\label{tb:subject_sas}
      \resizebox{\textwidth}{!}{
	\begin{threeparttable}  
		\setlength{\tabcolsep}{1mm}
\begin{tabular}{llllllrrrr}
\toprule
\textbf{LLM System} & \textbf{Domain} & \textbf{Task Dataset} & \textbf{$\#$Objectives} & \textbf{$\#$Cost} & \textbf{$\#$HP} & \textbf{$\#$$\boldsymbol{\mathcal{C}}$} & \textbf{$\#$$\boldsymbol{\mathcal{S}}$} & \textbf{$\#$$\boldsymbol{\mathcal{F}}$} & \textbf{$\#$$\boldsymbol{\mathcal{Z}}$} \\ \midrule
\textsc{LightRAG}~\cite{lightrag_github} & RAG & \textit{HotpotQA}~\cite{HotpotQA} & 9 & 10 & 23 & 1004-1025 & 23512 & 4 & 216 \\
\multirow{2}{*}{\textsc{NaiveRAG}~\cite{langchain_github}} & \multirow{2}{*}{RAG} & \textit{Agriculture}~\cite{lightrag}, \textit{Biography}~\cite{lightrag}, & \multirow{2}{*}{8} & \multirow{2}{*}{7} & \multirow{2}{*}{12} & \multirow{2}{*}{1000} & \multirow{2}{*}{56220} & \multirow{2}{*}{4} & \multirow{2}{*}{108} \\
 & & \textit{HotpotQA}~\cite{HotpotQA}, \textit{BioASQ}~\cite{BioASQ} &  &  &  &  &  &  &  \\
\textsc{HtmlRAG}~\cite{htmlrag_github} & RAG & \textit{Default-HTML Tasks}~\cite{DBLP:conf/www/TanDWWCW25} & 8 & 7 & 15 & 406-1000 & 27173 & 3 & 90 \\ 
\textsc{vLLM}~\cite{vllm_github} & Inference Engine & \textit{ShareGPT}~\cite{sharegpt_dataset} & 5 & 2 & 16 & 1070 & 115685 & 4 & 216 \\ 
\textsc{SGLang}~\cite{sglang_github} & Inference Engine & \textit{ShareGPT}~\cite{sharegpt_dataset} & 5 & 2 & 16 & 190 & 50849 & 5 & 108 \\ 
\textsc{AutoGPT}~\cite{autogpt_github} & Agent & \textit{AGBenchmark}~\cite{autogpt_github} & 5 & 3 & 12 & 92 & 9011 & 3 & 112 \\
\textsc{OpenHands}~\cite{openhands_github} & Agent & \textit{ProofWriter}~\cite{tafjord-etal-2021-proofwriter} & 3 & 6 & 21 & 1000 & 82000 & 4 & 82 \\
\bottomrule
\end{tabular}

          
		\footnotesize
        $\#$\textbf{Objective:} No. of objectives;
        $\#$\textbf{Cost:} No. of cost metrics;
		$\#$\textbf{HP:} No. of hyperparameter;
		$\#$\textbf{$\boldsymbol{\mathcal{C}}$:} No. of configurations per fidelity setting sampled so far;
        $\#$\textbf{$\boldsymbol{\mathcal{S}}$:} No. of total configuration samples collected so far;
        $\#$\textbf{$\boldsymbol{\mathcal{F}}$:} Dimension of fidelity factors;
		$\#$\textbf{$\boldsymbol{\mathcal{Z}}$:} No. of fidelity settings considered.
	\end{threeparttable}
      }
\end{table*}

\begin{enumerate}

    \item \textbf{Hyperparameter configuration space definition:} We define the hyperparameter configuration space by using all tunable hyperparameters for both the LLM and other system components from code interfaces, official documentation, and deployment settings. 
    

    \item \textbf{Fidelity space definition:} We set the fidelity space as a composition of controllable fidelity factors related to the standard and compatible task datasets (e.g., \textit{BioASQ}~\cite{BioASQ}), under which the LLM system is operating. Those factors should affect the measurement cost and objectives. Each factor has a range or set of values based on the system implementation, task design, and execution constraints. For instance, in \textsc{LightRAG}, the \texttt{context-count} specifies the number of contexts stored in the main retrieval corpus of the given task, where larger values increase the retrieval space and generally lead to more expensive measurements. 


    \item\textbf{Sampling strategy:} For the fidelity space, we first discretize each fidelity factor into a set of executable levels, then compose fidelity settings from their combinations, while excluding combinations that violate constraints or lead to invalid executions. Given the scale of configuration spaces, exhaustive measurement is impractical. We therefore adopt a sampling-based construction strategy, combining Latin hypercube sampling~\cite{mckay1992latin}, uniform random sampling, and distance-based sampling~\cite{kaltenecker2019distance} to improve coverage and representativeness under each fidelity setting. 

    \item\textbf{Execution and measurement pipeline:} For each sampled configuration-fidelity pair, we deploy the LLM system on a cluster of servers, each node with an Intel Core i9 24 cores CPU, RTX 5090 GPU (32GB VRAM, CUDA 13.0), and 128GB RAM. Every LLM system is profiled on a dedicated server to mitigate hardware noise. The underlying LLMs used are varied, such as \textsc{Llama-8B}, \textsc{Deepseek-r1-14B}, and \textsc{Qwen3-14B} etc.
    
    
    \item \textbf{Data validation and cleaning:} The collected raw records are subsequently validated and cleaned to ensure consistency, completeness, and usability. This process identifies failed or abnormal executions, filters invalid measurements, and standardizes the remaining records into a unified format for downstream analysis/usage.
\end{enumerate}

By May 2026, the dataset collected by \approach~has consumed $\approx2{,}277{,}600$ CPU core-hours (10 servers, 24 cores each) and $\approx94{,}900$ GPU-hours (10 servers). The details of LLM systems profiled so far have been summarized in Table~\ref{tb:subject_sas}, and more descriptions can be found in Appendix~\ref{app:systems}.

\subsection{Enclosed Folders and Datasets Format}
In \approach, the fidelity settings are organized as folders in the directory structure, as shown in Figure~\ref{fig:dirtree}. Here, \texttt{moderate-req1-code\_generation} is a setting with a total of three fidelity factors, where each item refers to the value of a factor separated by ``\texttt{-}''. This particular example means a fidelity setting of tasks with {moderate} difficulty under one request, and it belongs to a type of code generation task. There is a \texttt{.csv} file under each folder of the fidelity setting.

Every \texttt{.csv} file contains a standardized tabular format for storing measurements of configurations under a specific fidelity setting. Figure~\ref{fig:content} presents an example record for a system under one specific fidelity setting. As can be seen, the columns are (more details can be found in Appendix~\ref{app:metadata}):


\begin{figure}[t!]
    \centering
    \begin{minipage}{0.49\linewidth} 
        \scriptsize
        \dirtree{%
            .1 AutoGPT/.
            .2 moderate-req1-code\_generation/.
            .3 moderate-req1-code\_generation.csv.
            .3 hw\_file/.
            .4 hw-1.txt.
            .4 ....
            .3 log\_file/.
            .4 log-1.txt.
            .4 ....
            .2 simple-req1-code\_generation/.
            .2 simple-req2-logic\_puzzles/.
            .2 ....
        }
         \caption{Exampled directory structure for the LLM system \textsc{AutoGPT}.}
    \label{fig:dirtree}
    \end{minipage}  
    ~
        \begin{minipage}{0.49\linewidth} 
        \resizebox{\linewidth}{!}{
    

\setlength{\tabcolsep}{1.5mm}
\begin{tabular}{lc|lc|lc|ll}

\textbf{cfg-ai-max\_seqs} & ... & \textbf{obj-throughput$+$} & ... & \textbf{cost-duration($s$)} & ... & \textbf{hw-file}& \textbf{log-file} \\ \midrule

1024&...&145.2&...&84.7&...&hw-1.txt&log-1.txt\\
512&...&176.4&...&78.6&...&hw-2.txt&log-2.txt\\
256&...&156.2&...&83.4&...&hw-3.txt&log-3.txt\\
2048&...&198.4&...&67.5&...&hw-4.txt&log-4.txt\\
1024&...&242.5&...&56.8&...&hw-5.txt&log-5.txt\\
128&...&256.2&...&89.5&...&hw-6.txt&log-6.txt\\
1024&...&264.4&...&58.7&...&hw-7.txt&log-7.txt\\
64&...&175.3&...&74.3&...&hw-8.txt&log-8.txt\\
1024&...&146.6&...&65.8&...&hw-9.txt&log-9.txt\\
64&...&185.9&...&88.3&...&hw-10.txt&log-10.txt\\
1024&...&225.4&...&84.2&...&hw-11.txt&log-11.txt\\
512&...&245.7&...&82.6&...&hw-12.txt&log-12.txt\\
256&...&263.3&...&88.4&...&hw-13.txt&log-13.txt\\
1024&...&274.2&...&44.9&...&hw-14.txt&log-14.txt\\
512&...&178.8&...&64.7&...&hw-15.txt&log-15.txt\\
256&...&185.4&...&53.3&...&hw-16.txt&log-16.txt\\
1024&...&135.4&...&65.7&...&hw-17.txt&log-17.txt\\
32&...&195.3&...&66.8&...&hw-18.txt&log-18.txt\\
64&...&147.2&...&67.7&...&hw-19.txt&log-19.txt\\
1024&...&297.2&...&86.7&...&hw-20.txt&log-20.txt\\
...&...&...&...&...&...&...&...\\
512&...&265.4&...&85.6&...&hw-$n$.txt&log-$n$.txt\\


\end{tabular}

		}
         \caption{An example of the hyperparameter configuration data of \texttt{moderate-req1-code\_generation.csv}.}
   \label{fig:content}
    \end{minipage}
  
\end{figure}

\begin{itemize}
    \item \textbf{Hyperparameters} record all the tunable variables (with a prefix \texttt{cfg-ai-} and \texttt{cfg-} for AI and non-AI hyperparameters, respectively) of the target LLM system, with each column corresponding to one hyperparameter.
    \item \textbf{Objective metrics} denote the target task objectives of the evaluated LLM system during the inference stage, including both quality (e.g., F1-score) and performance (e.g., latency), with a prefix \texttt{obj-}. When applicable, the suffixes ``$+$'' and ``$-$'' indicate the metric to be maximized and minimized, respectively.
    \item \textbf{Cost metrics} include the measurement cost incurred by each configuration of the LLM systems, such as token usage and execution duration, indicated by a prefix \texttt{cost-}.
    \item \textbf{Hardware metrics} record additional hardware statistics observed during execution, such as CPU/GPU utilization and memory usage. This contains a link to the raw file (\texttt{hw-file}).
    \item \textbf{Logs} contain execution traces, such as error messages and other diagnostic signals that are essential for understanding system behaviors. This contains a link to the raw file (\texttt{log-file}).
\end{itemize}

\begin{figure}[t]
    \centering
\begin{lstlisting}[
style=pythonapicompact
]
from llmsys_hpobench import Benchmark

# initialize benchmark for one target LLM system
b = Benchmark(system="target_system")

# obtain searchable spaces
X = b.get_config_space()    # hyperparameter space
Z = b.get_fidelity_space()  # fidelity/environment space

# choose one fidelity/env setting
z = Z.sample()

# define optimization budget and consumption tracker
t = 0
budget = 3600

# iterative query under z
while t < budget:
    x = X.sample()
    M = b.evaluate(config=x, fidelity=z)  # M = f(x, z)

    perf = M["perf"]  # quality / latency / throughput / ...
    cost = M["cost"]  # token / time / resource cost / ...
    log = M["log"]    # trace / error / run metadata / ...
    t = t + cost

# users can select only needed fields for analysis
view = M.select(perf=[...], cost=[...], log=[...])
\end{lstlisting}
\caption{High-level pseudocode example of using the unified \approach\ API.}
\label{fig:api_pseudocode}
\end{figure}

\subsection{API Access of \approach}
\begin{figure*}[t]
    \centering
    \includegraphics[width=\linewidth]{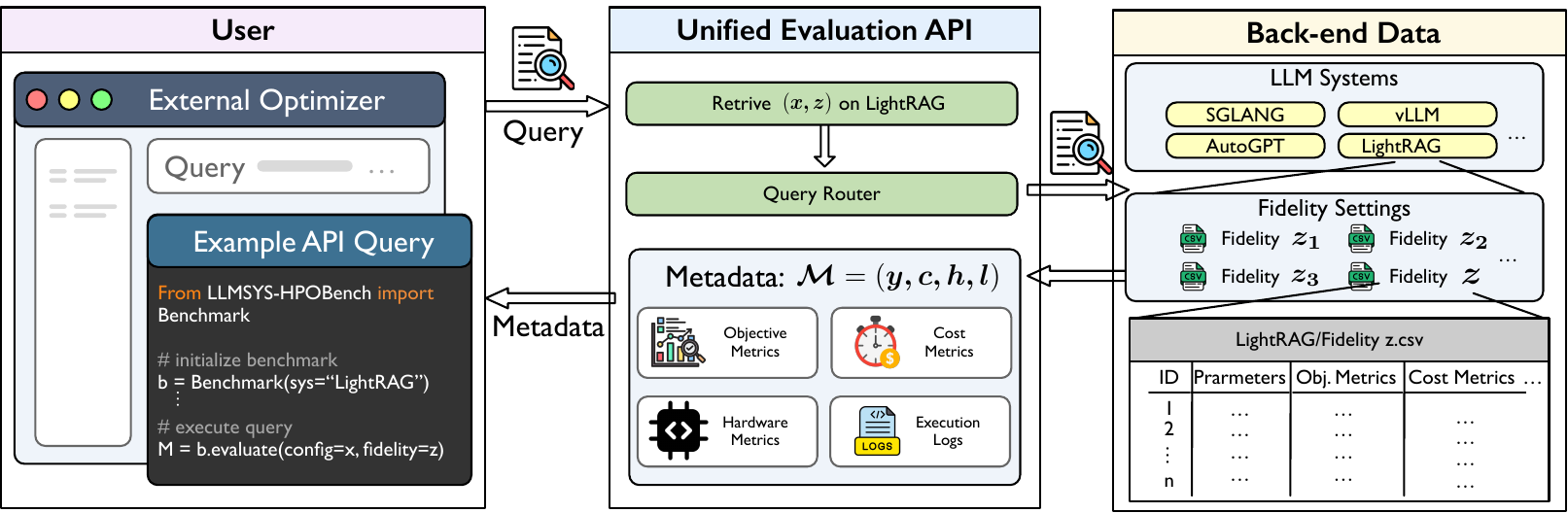}
    \caption{Overview of the \approach\ workflow.}
    \label{fig:API_usage}
\end{figure*}

To facilitate convenient and consistent use across different research scenarios, \approach\ is encapsulated as a unified benchmark abstraction with a standardized API. Figure~\ref{fig:api_pseudocode} is an example: in total of four lines of API calls, one can query the benchmark with a hyperparameter configuration $\bm{x}$ under a fidelity setting $\bm{z}$ and receive the structured metadata $\bm{\mathcal{M}}$, including all the column data shown in Figure~\ref{fig:content}. As for the workflow in Figure~\ref{fig:API_usage}, internally the benchmark is accessed through a unified interface that connects external users, such as HPO algorithms or analysis tools, with the back-end benchmark data. Given a hyperparameter configuration $\bm{x}$ and a fidelity setting $\bm{z}$, a user issues a query to the simple API in \approach. The query is then routed to the corresponding benchmark record associated with the target LLM system and fidelity setting, where the matched instance is retrieved from the back-end tabular data\footnote{Note that when a hyperparameter configuration queried cannot be found in the dataset, we return the objective value of its nearest neighbor, which is a pragmatic and widely-used mechanism~\cite{DBLP:conf/automl/PfistererSMBB22,DBLP:conf/iclr/ZelaSZLKH22,chen2026promisetune}.}. The API subsequently returns the structured metadata $\bm{\mathcal{M}}$. This API provides a unified way to query \approach\ and naturally supports a wide range of downstream tasks. Note that we intentionally avoid the use of surrogate models due to their unstable implications to the realism of optimization~\cite{DBLP:journals/tse/ChenGC25}.

\section{How to Use \approach? A Case Study}
\label{section:case_study}



\pgfplotsset{compat=1.18}

We now present a case study demonstrating how \approach\ can be used to run representative HPO algorithms under the same time-cost budget from the cost metrics. We consider \texttt{Random Search (RS)}, \texttt{Hyperband (HB)}~\cite{DBLP:journals/jmlr/LiJDRT17}, \texttt{BOHB}~\cite{DBLP:conf/icml/FalknerKH18}, \texttt{SMAC}~\cite{SMAC}, and \texttt{HEBO}~\cite{Cowen-Rivers2022-HEBO}, covering random exploration, multi-fidelity, and model-based black-box optimization.  More details can be found in Appendix~\ref{app:case}. The experiment is conducted on three representative systems from \approach, i.e., \textsc{NaiveRAG}, \textsc{LightRAG}, and \textsc{vLLM}, with one representative objective selected for each. Notably, we optimize lexical accuracy for \textsc{NaiveRAG} as it provides an aggregate measure of system-level accuracy over the entire evaluation set; F1 score for \textsc{LightRAG}, because it reflects end-to-end answer quality; and throughput for \textsc{vLLM}, since its primary goal is serving efficiency.

\begin{figure*}[t!]
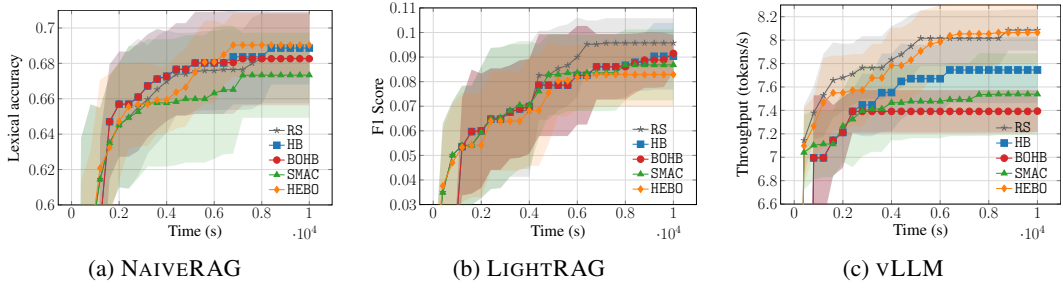

\centering

\begin{minipage}[t]{0.31\textwidth}
\centering
\subfloat[\textsc{NaiveRAG}]{
    \includestandalone[width=\linewidth]{figures/naiverag_error}
}
\end{minipage}
\hfill
\begin{minipage}[t]{0.31\textwidth}
\centering
\subfloat[\textsc{LightRAG}]{
    \includestandalone[width=\linewidth]{figures/lightrag_error}
}
\end{minipage}
\hfill
\begin{minipage}[t]{0.31\textwidth}
\centering
\subfloat[\textsc{vLLM}]{
    \includestandalone[width=\linewidth]{figures/vLLM_error}
}
\end{minipage}

\caption{Optimization trajectories of HPO algorithms over 10 runs using \approach. The x-axis is calculated via the cost metric ``\texttt{cost-runtime}''. \approach~enables the experiments to be completed in a few hours, which would otherwise require $\approx11$ days on real LLM systems.}
\label{fig:case}
\end{figure*}

To establish a unified evaluation protocol, we designate a relatively expensive fidelity setting for each studied LLM system as the target, under which the final performance of all algorithms is assessed. Single-fidelity algorithms (i.e., \texttt{RS}, \texttt{SMAC}, and \texttt{HEBO}) allocate their budget directly to target setting evaluations, whereas multi-fidelity HPO algorithms such as \texttt{HB} and \texttt{BOHB} can leverage lower-cost fidelity settings to guide the search before allocating evaluations to the target setting. This enables a budget-consistent comparison across different HPO algorithm families, while showcasing the ability of \approach\ to support fidelity-aware optimization under realistic time-cost constraints. As shown in Figure~\ref{fig:case}, the results have already revealed some new findings for LLM systems:


\begin{itemize}
    \item \textbf{Advanced HPO algorithms do not necessarily outperform simple baselines.} Their relative advantage varies markedly across LLM systems. On \textsc{NaiveRAG}, \texttt{HEBO} and \texttt{HB} achieve the strongest final performance, while \texttt{RS} is clearly weaker. In contrast, on both \textsc{LightRAG} and \textsc{vLLM}, \texttt{RS} remains highly competitive and in fact achieves the best results. This indicates that \approach\ induces heterogeneous optimization behaviors across LLM systems---the strong interactions cross AI and non-AI hyperparameters often makes non-trivial challenges for HPO algorithms (given \textbf{C1} and \textbf{C3}).
    \item \textbf{Multi-fidelity optimization is helpful, but its benefit is not universal.} On \textsc{NaiveRAG}, multi-fidelity methods, especially \texttt{HB}, deliver clear gains over \texttt{RS}, showing that lower-fidelity evaluations can provide useful guidance. However, this advantage does not carry over uniformly to \textsc{LightRAG} and \textsc{vLLM}, where single-fidelity baselines remain highly competitive. A likely reason is that the rich multi-factor fidelity space in \approach\ can induce non-monotonic and non-linear relationships between measurement cost and approximation of objective metrics in LLM systems, thereby posing a new challenge to existing multi-fidelity HPO methods (due to \textbf{C2}).
    \item \textbf{Algorithm superiority is budget-sensitive.} Under limited time-cost budgets, some algorithms reach strong solutions much faster, whereas with larger budgets, the advantage may shift to different algorithms for LLM systems. This highlights the need for cost-aware optimization therein. By recording the measurement cost of each queried configuration under rich multi-factor fidelity spaces, \approach\ naturally supports cost-aware analysis of HPO algorithms' behavior under explicit budget constraints (given \textbf{C5}).
\end{itemize}

Overall, this case study highlights the values of \approach\ as a benchmark for understanding HPO algorithms' behavior for LLM systems. Through jointly supporting real-world systems, heterogeneous hyperparameters, diverse objectives, explicit costs, and controllable fidelity settings, \approach\ naturally supports deeper analysis of their strengths, limitations, and applicability of HPO algorithms for LLM systems, pushing the boundary of the research field.

\usetikzlibrary{intersections}
\usepgfplotslibrary{fillbetween}

\section{Discussion}\label{section:discussion}



\subsection{AutoML Research Opportunities with \approach}
Beyond serving as a benchmark for algorithm comparison, \approach\ provides a reusable and convenient empirical resource for broader HPO research on LLM systems in the AutoML community. Below, we discuss these with respect to the characteristics of \approach.


\paragraph{{Short-term opportunities.}} Research directions that can be investigated in the near future include:

\begin{itemize}
    
\item \textbf{High-dimensional non-monotonic fidelity HPO} (using \textbf{C2}): \approach\ supports research on multi-fidelity HPO for LLM systems. In particular, the high dimensions and complexity of the fidelity factors can impose non-monotonic and non-linear relationships to the evaluation cost and objective metrics, which might invalidate the existing multi-fidelity HPO algorithms, e.g., \texttt{Hyperband}~\cite{DBLP:journals/jmlr/LiJDRT17}, \texttt{BOHB}~\cite{DBLP:conf/icml/FalknerKH18}, and \texttt{DEHB}~\cite{DBLP:conf/ijcai/AwadMH21}. All those raise the need for more specialized HPO algorithms.

\item \textbf{AI and non-AI interaction-aware HPO} (using \textbf{C3}): The unique understanding of AI and non-AI hyperparameters in \approach~creates opportunities and knowledge to tailor HPO algorithm, e.g., in RAG, studies have shown that the AI and non-AI hyperparameters can be optimized sequentially with discriminative interactions for better quality~\cite{DBLP:conf/ijcai/ChenCDS4RAG}. 


\item \textbf{Many-objective HPO} (using \textbf{C4}): Given the complexity of LLM systems, \approach~contains up to nine objective metrics to be simultaneously considered, which is often not the case for classic HPO scenarios. This additional high dimensionality, together with their diverse requirements~\cite{DBLP:conf/icse/WangChen26,DBLP:conf/icse/WangChenACL26,DBLP:journals/tosem/ChenL23a}, is known to be challenging~\cite{DBLP:journals/tse/LiCY22,DBLP:conf/cec/IshibuchiTN08,deb2013evolutionary}.

\item \textbf{Cost- and hardware-sensitive HPO} (using \textbf{C5} and \textbf{C6}): The fine-grained data of measurement cost and hardware consumption for individual hyperparameter configuration in \approach~brings the unique chance to incorporate that information in HPO. This is particularly true when the different cost metrics can vary considerably, for example, varying RAG hyperparameters like retrieval depth can double end-to-end token costs and duration of measurements~\cite{DBLP:journals/corr/abs-2502-18635}. The same also applies to the hardware metrics, which are fundamental for AI inference, as different batching or KV-cache configurations can drastically shift GPU/memory consumption~\cite{bambhaniya2026mistcodesignframeworkheterogeneous}. Indeed, cost- and hardware-sensitive HPO can be explored by additionally discriminating hyperparameter configurations on their multi-dimensional cost and hardware measurements. This is especially useful under multi-fidelity settings.


\end{itemize}

\paragraph{{Intermediate-term opportunities.}} Research problems that require some effort to explore/define:

\begin{itemize}
\item \textbf{Multi-modal LLM-based HPO algorithm creation} (using \textbf{C7}): \approach~contains the execution logs from the LLM systems, which can be used as part of the modality of text prompt and optimization data to instruct a LLM for automatically creating new HPO algorithms. This is also an active topic for automated general algorithm creations~\cite{DBLP:conf/aaai/LiuLZTY26}.

\item \textbf{Explainable HPO} (using \textbf{C7}): The log information in \approach~enables one to understand the internal working mechanism of the LLM system that goes beyond the individual LLM and/or the typical system components (e.g., a tool or vector database). This also provides opportunities for the explainability of HPO behaviors.

\end{itemize}


\paragraph{{Long-term opportunities.}} Directions that have yet reached any established foundation are:

\begin{itemize}
\item \textbf{Empirical study of HPO via fitness landscape analysis} (using \textbf{C1}--\textbf{C4}): With the large number of cases/environments available in \approach, one can conduct multiple empirical studies to understand the domain knowledge of LLM systems via fitness landscape analysis~\cite{pitzer2012comprehensive,10.1145/3803859}. This can stimulate new directions of landscape-driven HPO.

\item \textbf{Industry and real-world uptakes} (using \textbf{C1}): The real-world property of \approach~ensures that any HPO algorithms built on \approach~can be applied directly to real-world LLM systems without many barriers. 
\end{itemize}

\subsection{\approach~as a Live Benchmark Suite}
Given the easy-to-extend portal available in \approach, it allows further extension by designs. In fact, as a live benchmark, it is growing in phases at two aspects (see Appendix~\ref{app:lm}):
\begin{itemize}
    \item More hyperparameter configuration data, hardware data, and logs of the LLM systems.
    \item Data for more LLM systems will become available.
    
\end{itemize}
Of course, in addition to the efforts from the authors, we also invite the community to contribute by providing new data, sources, or recommending LLM systems to be profiled.


\section{Conclusion}
\label{sectioin:conclusion}

In this paper, we present \approach, the first open-sourced live benchmark for HPO of LLM systems. \approach~is unique in the sense that it contains a mix of hyperparameter configuration space of both AI (e.g., LLM) and non-AI components (e.g., memory or vector database) under varying fidelity factors that exhibit non-monotonic and nonlinear relationships to cost and inference objectives, together with measurement cost of individual configuration, hardware metrics, and logs. We release \approach~as a readily usable benchmark suite, hoping that it would play a non-trivial role in shaping the future research of AutoML in the LLM era.

\bibliographystyle{plain}
\bibliography{references}

\newpage
\appendix

\section{Limitations}
\label{app:limitations}

\approach\ is designed to provide a realistic and extensible benchmark for LLM systems HPO, but its current release inevitably reflects several construction trade-offs. The main limitations are summarized below.

\begin{itemize}
    \item \textbf{LLM system coverage.} \approach\ currently has yet to cover a large number of LLM systems. That said, as a live benchmark, \approach~will be further extended periodically to include more real-world LLM systems.
    \item \textbf{Fidelity space coverage.} To make benchmark construction executable, the fidelity space of \approach\ is instantiated through a reduced, finite set of discretized settings. Yet, still, it goes well beyond prior benchmarks in fidelity, richness/coverage, and contains the most common settings in practice.
    \item \textbf{Hyperparameter space coverage.} The hyperparameter configurations in \approach\ are constructed through sampling, which does not cover the full space. Yet, since exhaustive measurements are impractical for real-world LLM systems at scale, the ensembles of sampling methods used therein are common practice to ensure representativeness. Further, given the intuitive/easy-to-maintain structure in \approach, more data can be easily appended in phases.
    \item \textbf{Measurement noises.} Time-related metrics would inevitably be subject to noise from the infrastructure and hardware. Yet, our data cleaning and curation process has ensured that the abnormal samples are removed, and that the measurements are conducted repeatably to preserve reliability. 
    
\end{itemize}

\begin{table}[h!]
	\centering
	\caption{Comparing \approach~with existing HPO and system benchmarks.}
    \label{tb:compare-benchmarks}
      \resizebox{\textwidth}{!}{
	\begin{threeparttable}  
		\setlength{\tabcolsep}{1mm}
\begin{tabular}{lllllllll}
\toprule
\textbf{Benchmark} & \textbf{Domain} & \textbf{$\#$HP (AI)} & \textbf{$\#$HP (non-AI)} & \textbf{$\#$$\boldsymbol{\mathcal{F}}$}  & \textbf{$\#$Objectives} & \textbf{$\#$Cost} & \textbf{Hardware Metrics?} & \textbf{Logs?} \\ \midrule

\texttt{NAS-Bench-101}~\cite{DBLP:conf/icml/YingKCR0H19} & NAS & 26 & 0 & 1  & 3 & 1 & \textcolor{red}{\ding{55}} & \textcolor{red}{\ding{55}}\\

\texttt{NAS-Bench-201}~\cite{DBLP:conf/iclr/Dong020} & NAS & 6 & 0 & 1  & 6 & 2 & \textcolor{red}{\ding{55}} & \textcolor{red}{\ding{55}}\\

\texttt{HPO-B}~\cite{DBLP:conf/nips/Pineda-ArangoJW21} & HP & 1-53 & 0 & \textcolor{red}{\ding{55}}  & 1 & \textcolor{red}{\ding{55}} & \textcolor{red}{\ding{55}} & \textcolor{red}{\ding{55}}\\

\texttt{HPOBench}~\cite{eggensperger2021hpobench} & HP/NAS & 2-26 & 0 & 1-2  & 1-4 & 2 & \textcolor{red}{\ding{55}} & \textcolor{red}{\ding{55}}\\ 

\texttt{YAHPO Gym}~\cite{DBLP:conf/automl/PfistererSMBB22} & HP & 2-38 & 0 & 1  & 2-12 & 1-5 & \textcolor{red}{\ding{55}} & \textcolor{red}{\ding{55}}\\ 

\texttt{JAHS-Bench-201}~\cite{DBLP:conf/nips/BansalSJZH22} & HP/NAS & 10 & 0 & 4  & 10 & 4 & \textcolor{red}{\ding{55}} & \textcolor{red}{\ding{55}}\\ \midrule

Mühlbauer et al.~\cite{muhlbauer2023analysing} & System & 0 & 11-33 & 1  & 1 & \textcolor{red}{\ding{55}} & \textcolor{red}{\ding{55}} & \textcolor{red}{\ding{55}}\\

Nair et al. \cite{DBLP:journals/tse/Nair0MSA20} & System & 0 & 3-39 & \textcolor{red}{\ding{55}}  & 1-2 & \textcolor{red}{\ding{55}} & \textcolor{red}{\ding{55}} & \textcolor{red}{\ding{55}}\\

Krishna et al. \cite{DBLP:journals/tse/KrishnaNJM21} & System & 0 & 12-57 & 1  & 1 & \textcolor{red}{\ding{55}} & \textcolor{red}{\ding{55}} & \textcolor{red}{\ding{55}}\\

Weber et al. \cite{DBLP:conf/icse/WeberKSAS23} & System & 0 & 2-36 & \textcolor{red}{\ding{55}}  & 2-4 & 1 & \textcolor{red}{\ding{55}} & \textcolor{red}{\ding{55}}\\

MOOT$^*$ \cite{DBLP:journals/corr/abs-2511-16882} & System & 0 & 3-39 & 1  & 1-3 & \textcolor{red}{\ding{55}} & \textcolor{red}{\ding{55}} & \textcolor{red}{\ding{55}}\\ \midrule

\approach~(ours) & LLM system & 5-8 & 4-17 & 3-5   & 3-9 & 2-10 & \textcolor{teal}{\ding{51}} & \textcolor{teal}{\ding{51}}\\ \bottomrule

\end{tabular}          
		\footnotesize
        $\#$\textbf{Objectives:} No. of objectives;
		$\#$\textbf{HP:} No. of hyperparameters;
        $\#$\textbf{$\boldsymbol{\mathcal{F}}$:} Dimension of fidelity factors.
        $\#$\textbf{Cost:} No. of cost metrics.
	\end{threeparttable}
      }
\end{table}

\section{Detailed Comparisons to Existing Benchmarks}
\label{app:benchmark_comparison}

Table~\ref{tb:compare-benchmarks} provides an aspect-wise comparison between \approach\ and representative HPO/system benchmarks. The main differences can be summarized as follows:

\begin{itemize}
    \item \textbf{Targeted systems/models.} Unlike conventional HPO benchmarks for ML/NAS and system benchmarks for traditional software systems, \approach\ focuses on real-world end-to-end LLM systems, which is a synergy of both.
    \item \textbf{Hyperparameters.} Prior benchmarks typically expose either AI or non-AI hyperparameters only, but not both; \approach\ jointly covers both, with heterogeneous spaces spanning 5--8 AI hyperparameters and 4--17 non-AI hyperparameters, respectively.
    \item \textbf{Fidelity factors.} Existing HPO benchmarks typically support one to two fidelity factors, often instantiated through resource variables such as epochs or data fraction. System benchmarks usually provide no explicit fidelity abstraction, or only a single workload-related factor. By comparison, \approach\ exposes richer fidelity settings, with 3--5 dimensional fidelity spaces.
    \item \textbf{Objective metrics.} Existing HPO and system benchmarks often contain only a limited set of objective metrics, while \approach\ records up to nine objective metrics, enabling an in-depth analysis of trade-offs in LLM systems HPO. Only two existing benchmarks consider higher dimensions of objectives than \approach.
    \item \textbf{Cost metrics.} Prior benchmarks generally provide sparse or no explicit cost metrics. By comparison, \approach\ records up to 10 cost metrics, which supports more systematic study of cost-aware optimization.
    \item \textbf{Hardware metrics.} Hardware metrics are rarely included in existing HPO or system benchmarks. \approach\ explicitly records hardware metrics for each evaluation, providing an additional view of execution behavior beyond objective and cost metrics.
    \item \textbf{Logs.} Execution logs are typically unavailable in prior benchmarks. \approach\ well records logs for each evaluation, supporting inspection and explainable analysis. 
\end{itemize}


\section{Detailed Setup of the Case Study}
\label{app:case}

The case study is conducted on a standard machine with an Intel Core i9-13900H CPU with 14 cores, 20 threads and 32 GB RAM. For an experiment run that requires at least 110 minutes for a HPO algorithm to run on the real LLM system, using \approach~merely needs a few seconds to several tens of seconds to complete. This significantly reduces the efforts required and hence expedite the research progress/discovery in the field.

The considered HPO algorithms in the case study are specified below. Note that for all algorithms, we use their default hyperparameter values.


\begin{itemize}
    \item \textbf{{Random Search (RS)}:} \texttt{RS} is a pure exploration optimizer that samples configurations randomly from the hyperparameter configuration space.
    \item \textbf{{Hyperband (HB)}~\cite{DBLP:journals/jmlr/LiJDRT17}:} \texttt{HB} is a multi-fidelity optimizer that improves search efficiency via adaptive resource allocation. It measures many configurations under low-fidelity settings with small budgets, prunes poorly performing candidates early, and progressively allocates more resources to promising configurations.
    \item \textbf{{BOHB}~\cite{DBLP:conf/icml/FalknerKH18}:} \texttt{BOHB} combines Bayesian optimization (BO) with \texttt{HB}-style resource allocation for multi-fidelity optimization. It preserves the early-stopping mechanism of \texttt{HB} while using a probabilistic model to guide the sampling of promising configurations, thereby improving both sample efficiency and budget utilization.
    \item {\textbf{SMAC}~\cite{SMAC}:} \texttt{SMAC} is a sequential model-based optimizer that uses a surrogate model, typically based on random forests, to estimate configuration performance. It selects new configurations by optimizing an acquisition function, making it well-suited for expensive and mixed-type hyperparameter optimization.
    \item {\textbf{HEBO}~\cite{Cowen-Rivers2022-HEBO}:} \texttt{HEBO} is a robust BO method for noisy and heterogeneous search spaces. It improves standard BO with input/output transformations, heteroscedastic uncertainty modeling, and evolutionary acquisition optimization, making it effective for mixed-type hyperparameter configuration spaces.
\end{itemize}

\section{More Details of \approach}
\label{app:more_details}

\subsection{LLM Systems}
\label{app:systems}


Here, we list a brief description of the LLM systems in \approach. However, it is worth noting that since \approach~is a live benchmark, more LLM systems will be added progressively.

\begin{itemize}
    \item \textbf{\textsc{LightRAG}:}
    \textsc{LightRAG} is a graph-enhanced retrieval-augmented generation framework that integrates knowledge graph-based indexing with a dual-level retrieval paradigm. 
    It enables efficient and context-aware information retrieval by jointly leveraging entity-level and global semantic search, while supporting incremental updates for low-cost and adaptive knowledge integration.
    \item \textbf{\textsc{NaiveRAG}:}
    \textsc{NaiveRAG} is a modular retrieval-augmented generation pipeline that follows the standard chunk-retrieve-generate paradigm.
    It constructs configurable vector indexes over chunked corpora and performs top-$k$ similarity retrieval to provide evidence for downstream answer generation.

    \item \textbf{\textsc{HtmlRAG}:}
    \textsc{HtmlRAG} is an HTML-focused retrieval-augmented generation framework tailored to semi-structured web documents.
    It first compresses noisy DOM content through a two-stage pruning process, combining embedding-based filtering with generation-model-based refinement, and then retrieves task-relevant evidence for downstream question answering.

    \item \textbf{\textsc{vLLM}:}
    \textsc{vLLM} is a high-throughput LLM inference engine designed around efficient batching and KV-cache management, providing an OpenAI-compatible serving API. It targets low-latency, high-concurrency generation with configurable scheduling and memory policies.
    \item \textbf{\textsc{SGLang}:}
    \textsc{SGLang} is a system for structured LLM serving and orchestration that emphasizes scheduling and memory optimization for long-context and multi-request workloads. It supports fine-grained control of cache, prefill, and parallelism to balance throughput and latency.
    \item \textbf{\textsc{AutoGPT}:}
    \textsc{AutoGPT} is an autonomous agent framework that couples LLM reasoning with tool use, memory, and execution loops. It is a representative of agentic AI systems that solve multi-step tasks through iterative planning and action.
    \item \textbf{\textsc{OpenHands}:}
    \textsc{OpenHands} is an open-source, model-agnostic platform for autonomous software development agents. 
    It supports code-related task execution through sandboxed runtimes, tool use, and SDK/API interfaces, making it a representative agentic coding system in \approach.
\end{itemize}



\subsection{Maintenance and Dissemination}
\label{app:lm}







\paragraph{Under what license the \approach~is released?} \approach~is released under the GNU General Public License v3.0. 

\paragraph{Who is maintaining the benchmark?} The maintenance and consolidation of \approach~will be completed by the authors' research group.


\paragraph{How are the updates conducted?} The update to \approach~will be conducted in phases; the current plan is to update it quarterly as new data become available. The items to be updated include both the new LLM systems and new data for the existing LLM systems.

\paragraph{How will the datasets in the \approach~be distributed?} The entire volume of data in \approach~can potentially become very large (i.e., currently the size of all file is approximately 38.5GB); however, we have designed a multiple-level chunking mechanism that can ensure users only need to download the parts that one really needs. In all cases, the API would remain functional. For example, the datasets are split into different LLM systems at different fidelity settings. The hardware and log files are also separated from the \texttt{.csv} data---if one is only interested in testing the objective values of hyperparameter configurations, s/he could only need to download the   \texttt{.csv} files.

\paragraph{How to ask questions or raise issues?} One can use the issue tracking service of our repository, which is a very typical feature of popular hosting platforms like GitHub.

\paragraph{How to contribute to \approach?} We have provided a comprehensive, intuitive, yet straightforward step-by-step guideline on how to contribute new data for \approach:


\begin{enumerate}
    \item \textbf{Open a tracking issue.} The contributor first describes the target LLM system, its family (RAG pipeline, inference engine, or agentic system), the intended contribution type (new system, new fidelity settings, or additional measurements), and the expected hardware/software requirements.
    \item \textbf{Create the system manual.} Add a manual under the corresponding documentation directory, i.e., \texttt{RAG/manuals/}, \texttt{Engine/manuals/}, or \texttt{Agent/manuals/}. The manual should define the AI and non-AI hyperparameters, their types and value ranges, implementation references, default settings, and any deployment assumptions.
    \item \textbf{Implement the benchmark interface.} Add scripts, samplers, clients, or workload generators under the matching code directory. Each contribution should expose the same conceptual interface as existing entries in \approach: a configuration space, a fidelity space, a runnable evaluation procedure, and machine-readable metadata for the returned objectives, costs, hardware traces, and logs.
    \item \textbf{Specify fidelities and sampling policy.} Document all fidelity dimensions, their allowed values, the Cartesian product or sampling strategy, and their expected impact on cost/resources. If the full Cartesian product is too expensive, the contributor should justify the reduced sampling plan and include a small pre-experiment showing that the selected fidelities remain informative.
    \item \textbf{Generate and organize measurements.} Store raw outputs, processed \texttt{.csv} summaries, cost logs, error logs, hardware traces, and visualizations in the standardized data layout. Every measured configuration should have a stable identifier so that objectives, costs, hardware metrics, and execution logs can be joined unambiguously.
    \item \textbf{Record the environment.} Add a reproducibility note with hardware specifications, operating system, Python/CUDA/library versions, launch commands, random seeds, container or virtual-environment details, and known failure modes.
    \item \textbf{Run validation before review.} Execute a smoke test on a small configuration--fidelity subset, check that the processed files match the metadata schema of \approach, verify that failed or partial runs are explicitly marked, and confirm that the provided scripts can reproduce at least one sample end-to-end.
    \item \textbf{Submit a pull request.} The pull request should link the tracking issue, summarize the added system/fidelity/data volume, list the validation result, and identify any limitations. Reviewers then check the manual, benchmark interface, data schema, and reproducibility artifacts before merging.
\end{enumerate}

More details can be found at here: \texttt{\textcolor{blue}{\url{https://github.com/ideas-labo/llmsys-hpobench/blob/main/CONTRIBUTING.md}}}.

\subsection{Full Metadata}
\label{app:metadata}


We showcase the details of metadata on hyperparameter space in Tables~\ref{tab:lightrag-knobs}--\ref{tab:openhands_knobs_table}; fidelity space in Table~\ref{tb:fidelity_space}; the objective metrics in Table~\ref{tb:obj}; the cost metrics in Table~\ref{tb:cost}; the hardware information in Table~\ref{tb:hardware}; and an example of the logging trace file in Figure~\ref{fig:autogpt-log}.



\begin{table}[h!]
    \centering
    \caption{Details of hyperparameters of \textsc{LightRAG}.}
    \begin{adjustbox}{width=\linewidth,center}
	\begin{threeparttable}
		\begin{tabular}{lllll}
\toprule
\textbf{Configuration Option} & \textbf{Component} & \textbf{Type} & \textbf{Default} & \textbf{Range} \\ \midrule

\texttt{llm\_model\_name} 
& AI 
& Categorical 
& llama3.1:8b 
& \makecell[l]{\{llama3.1:8b, qwen3:8b, deepseek-r1:14b,\\ 
gemma3:12b, qwen3:14b\}} \\
\texttt{llm\_model\_temperature} & AI & Float & 0.7 & [0.2, 1.2] \\
\texttt{llm\_model\_top\_p} & AI & Float & 0.9 & [0.0, 1.0] \\
\texttt{llm\_model\_top\_k} & AI & Integer & 40 & [5, 80] \\
\texttt{llm\_model\_repeat\_penalty} & AI & Float & 1.1 & [0.9, 1.5] \\
\texttt{llm\_model\_num\_ctx} & AI & Integer & 32768 & [20480, 40480] \\ \midrule
\texttt{max\_token\_for\_text\_unit} & non-AI & Integer & 4000 & [2000, 6000] \\
\texttt{max\_token\_for\_global\_context} & non-AI & Integer & 4000 & [2000, 6000] \\
\texttt{max\_token\_for\_local\_context} & non-AI & Integer & 4000 & [2000, 6000] \\
\texttt{llm\_model\_max\_token\_size} & non-AI & Integer & 8192 & [8192, 32768] \\

\texttt{chunk\_token\_size} & non-AI & Integer & 1200 & [800, 2000] \\
\texttt{chunk\_overlap\_token\_size} & non-AI & Integer & 120 & [50, 200] \\
\texttt{entity\_extract\_max\_gleaning} & non-AI & Integer & 1 & [1, 4] \\
\texttt{embedding\_batch\_num} & non-AI & Integer & 32 & [16, 64] \\
\texttt{embedding\_func\_max\_async} & non-AI & Integer & 16 & [8, 32] \\
\texttt{force\_llm\_summary\_on\_merge} & non-AI & Integer & 6 & [3, 15] \\
\texttt{llm\_model\_max\_async} & non-AI & Integer & 5 & [3, 10] \\
\texttt{max\_parallel\_insert} & non-AI & Integer & 5 & [3, 15] \\
\texttt{summary\_to\_max\_tokens} & non-AI & Integer & 500 & [250, 1000] \\
\texttt{mode} & non-AI & Categorical & global & \{local, global, hybrid, naive\} \\

\texttt{response\_type} 
& non-AI 
& Categorical 
& Multiple Paragraphs
& \makecell[l]{\{Multiple Paragraphs, \\Single Paragraph, Bullet Points\}} \\

\texttt{top\_k} & non-AI & Integer & 60 & [30, 100] \\
\texttt{cosine\_threshold} & non-AI & Float & 0.2 & [0, 0.6] \\

\bottomrule
\end{tabular}
	\end{threeparttable}
    \end{adjustbox}
    \label{tab:lightrag-knobs}
\end{table}

\begin{table}[h!]
    \centering
    \small
    \setlength{\tabcolsep}{18pt}
    \caption{Details of hyperparameters of \textsc{NaiveRAG}.}
    \begin{adjustbox}{width=\linewidth,center}
	\begin{threeparttable}
		\begin{tabular}{lllll}
\toprule
\textbf{Configuration Option} & \textbf{Component} & \textbf{Type} & \textbf{Default} & \textbf{Range} \\ 
\midrule
\texttt{embedding\_temperature} & AI & Float & 0.8 & [0, 1.0] \\
\texttt{embedding\_num\_ctx} & AI & Integer & 1024 & [512, 2047] \\
\texttt{embedding\_top\_k} & AI & Integer & 50 & [10, 99] \\
\texttt{embedding\_repeat\_penalty} & AI & Float & 1.1 & [0.9, 1.5] \\
\texttt{chat\_temperature} & AI & Float & 0.8 & [0, 1.0] \\
\texttt{chat\_num\_ctx} & AI & Integer & 4096 & [512, 8191] \\
\texttt{chat\_top\_k} & AI & Integer & 50 & [10, 99] \\
\texttt{chat\_repeat\_penalty} & AI & Float & 1.1 & [0.9, 1.5] \\ \midrule

\texttt{database\_type} & non-AI & Categorical & 1 & \{1, 2, 3\} \\
\texttt{chunk\_size} & non-AI & Integer & 512 & [256, 2047] \\
\texttt{chunk\_overlap} & non-AI & Integer & 64 & [32, 255] \\
\texttt{retriever\_k} & non-AI & Integer & 5 & [1, 9] \\
\bottomrule
\end{tabular}
	\end{threeparttable}
    \end{adjustbox}
    \label{tab:naiverag-knobs}
\end{table}

\begin{table}[h!]
    \centering
    \small
    \setlength{\tabcolsep}{18pt}
    \caption{Details of hyperparameters of \textsc{HtmlRAG}.}
    \begin{adjustbox}{width=\linewidth,center}
	\begin{threeparttable}
		\begin{tabular}{lllll}
\toprule
\textbf{Configuration Option} & \textbf{Component} & \textbf{Type} & \textbf{Default} & \textbf{Range} \\ 
\midrule
\texttt{embedding\_temperature} & AI & Float & 0.8 & [0.6, 1.0] \\
\texttt{embedding\_num\_ctx} & AI & Integer & 2048 & [1024, 4095] \\
\texttt{embedding\_top\_k} & AI & Integer & 20 & [0, 49] \\
\texttt{embedding\_repeat\_penalty} & AI & Float & 1.1 & [0.9, 1.3] \\
\texttt{chat\_temperature} & AI & Float & 0.4 & [0.2, 0.6] \\
\texttt{chat\_num\_ctx} & AI & Integer & 4096 & [2048, 8191] \\
\texttt{chat\_top\_k} & AI & Integer & 40 & [20, 59] \\
\texttt{chat\_repeat\_penalty} & AI & Float & 1.2 & [1.0, 1.4] \\ \midrule

\texttt{chunk\_size} & non-AI & Integer & 512 & [256, 1023] \\
\texttt{chunk\_overlap} & non-AI & Integer & 64 & [32, 127] \\
\texttt{retriever\_k} & non-AI & Integer & 5 & [3, 9] \\
\texttt{max\_node\_words\_embed} & non-AI & Integer & 64 & [32, 127] \\
\texttt{max\_context\_window\_embed} & non-AI & Integer & 512 & [256, 1023] \\
\texttt{max\_node\_words\_gen} & non-AI & Integer & 32 & [16, 63] \\
\texttt{max\_context\_window\_gen} & non-AI & Integer & 512 & [256, 1023] \\
\bottomrule
\end{tabular}
	\end{threeparttable}
    \end{adjustbox}
    \label{tab:htmlrag-knobs}
\end{table}


\begin{table}[h!]
    \centering
    \small
    \setlength{\tabcolsep}{18pt}
    \caption{Details of hyperparameters of \textsc{vLLM}.}
    \begin{adjustbox}{width=\linewidth,center}
	\begin{threeparttable}
		\begin{tabular}{lllll}
\toprule
\textbf{Configuration Option} & \textbf{Component} & \textbf{Type} & \textbf{Default} & \textbf{Range} \\
\midrule
\texttt{model} & AI & String & Llama-3.1-8B & model/repo id \\
\texttt{draft\_model} & AI & String & None & model/repo id \\
\texttt{num\_speculative\_tokens} & AI & Integer & 5 & [3, 10] \\
\texttt{top\_k} & AI & Integer & -1 & [10, 100] \\
\texttt{min\_p} & AI & Float & 0.0 & [0.8, 0.95] \\
\texttt{repetition\_penalty} & AI & Float & 1.0 & [1.1, 1.3] \\
\texttt{length\_penalty} & AI & Float & 1.0 & [0.8, 1.2] \\
\texttt{best\_of} & AI & Integer/None & None & \{None\} or [2, 5] \\ \midrule
\texttt{tensor\_parallel\_size} & non-AI & Integer & 1 & [1, \#GPUs] \\
\texttt{max\_num\_seqs} & non-AI & Integer & 256 & [64, 8192] \\
\texttt{max\_num\_batched\_tokens} & non-AI & Integer & 2048 & [64, 8192] \\
\texttt{block\_size} & non-AI & Categorical & 16 & \{8, 16, 32\} \\
\texttt{scheduler\_delay\_factor} & non-AI & Float & 0.0 & [0, 2] \\
\texttt{enable\_chunked\_prefill} & non-AI & Boolean & False & \{True, False\} \\
\texttt{enable\_prefix\_caching} & non-AI & Boolean & True & \{True, False\} \\
\texttt{disable\_custom\_all\_reduce} & non-AI & Boolean & False & \{True, False\} \\
\texttt{use\_v2\_block\_manager} & non-AI & Boolean & True & \{True, False\} \\
\bottomrule
\end{tabular}

	\end{threeparttable}
    \end{adjustbox}
    \label{tab:vllm-knobs}
\end{table}

\begin{table}[h!]
    \centering
    \small
    \setlength{\tabcolsep}{18pt}
    \caption{Details of hyperparameters of \textsc{SGLang}.}
    \begin{adjustbox}{width=\linewidth,center}
	\begin{threeparttable}
		\begin{tabular}{lllll}
\toprule
\textbf{Configuration Option} & \textbf{Component} & \textbf{Type} & \textbf{Default} & \textbf{Range} \\
\midrule
\texttt{top\_k} & AI & Integer & -1 & [10, 100] \\
\texttt{min\_p} & AI & Float & 0.0 & [0.8, 0.95] \\
\texttt{repetition\_penalty} & AI & Float & 1.0 & [1.1, 1.3] \\
\texttt{length\_penalty} & AI & Float & 1.0 & [0.8, 1.2] \\
\texttt{best\_of} & AI & Integer/None & None & \{None\} or [2, 5] \\ \midrule
\texttt{context\_length} & non-AI & Integer & model max & [1, model max] \\
\texttt{mem\_fraction\_static} & non-AI & Float & 0.8 & (0, 1] \\
\texttt{max\_total\_tokens} & non-AI & Integer & auto & [1, +inf) \\
\texttt{chunked\_prefill\_size} & non-AI & Integer & -1 & \{-1\} or [1, +inf) \\
\texttt{schedule\_policy} & non-AI & Categorical & fcfs & \{fcfs, priority\} \\
\texttt{dtype} & non-AI & Categorical & auto & \{auto, float16, bfloat16, float32\} \\
\texttt{quantization} & non-AI & Categorical & none & \{none, fp8, int8, int4\} \\
\texttt{kv\_cache\_dtype} & non-AI & Categorical & auto & \{auto, fp8\_e5m2, fp8\_e4m3\} \\
\texttt{speculative\_algorithm} & non-AI & Categorical & none & \{none, eagle, draft\} \\
\texttt{speculative\_num\_steps} & non-AI & Integer & 3 & [1, +inf) \\
\texttt{enable\_metrics} & non-AI & Boolean & True & \{True, False\} \\
\texttt{log\_level} & non-AI & Categorical & info & \{debug, info, warning\} \\

\bottomrule
\end{tabular}

	\end{threeparttable}
    \end{adjustbox}
    \label{tab:sglang-knobs}
\end{table}

\begin{table}[h!]
    \centering
    \small
    \setlength{\tabcolsep}{18pt}
    \caption{Details of hyperparameters of \textsc{AutoGPT}.}
    \begin{adjustbox}{width=\linewidth,center}
	\begin{threeparttable}
		\begin{tabular}{lllll}
\toprule
\textbf{Configuration Option} & \textbf{Component} & \textbf{Type} & \textbf{Default} & \textbf{Range} \\
\midrule
\texttt{fast\_llm} & AI & Categorical & Qwen2.5-1.5B & \{none, Qwen2.5-1.5B, Qwen2.5-3B\} \\
\texttt{smart\_llm} & AI & Categorical & Qwen2.5-7B & \{Qwen2.5-7B\} \\
\texttt{big\_brain} & AI & Boolean & \texttt{True} & \{\texttt{True}, \texttt{False}\} \\
\texttt{send\_token\_limit} & AI & Integer/None & \texttt{None} & \{\texttt{None}, 2048\} \\
\texttt{llm\_temperature} & AI & Float & 0.0 & \{0.0, 0.7\} \\
\texttt{llm\_max\_output\_tokens} & AI & Integer & 512 & \{512, 2048\} \\
\texttt{use\_functions\_api} & AI & Boolean & \texttt{False} & \{\texttt{True}, \texttt{False}\} \\
\midrule
\texttt{enabled\_components} & non-AI & Categorical & all enabled & \{FM, FM+CE, FM+CE+WS, FM+CE+WS+CTX\} \\
\texttt{allow\_fs\_access} & non-AI & Boolean & \texttt{False} & \{\texttt{True}, \texttt{False}\} \\
\texttt{full\_message\_count} & non-AI & Integer & 4 & \{2, 4, 8\} \\
\texttt{shell\_command\_control} & non-AI & Categorical & allowlist & \{allowlist, denylist\} \\
\texttt{cycle\_budget} & non-AI & Integer/None & 3 & \{3, 10, 25\} \\
\bottomrule
\end{tabular}

	\end{threeparttable}
    \end{adjustbox}
    \label{tab:autogpt-knobs}
\end{table}

\begin{table}[h!]
    \centering
    \small
    \setlength{\tabcolsep}{18pt}
    \caption{Details of hyperparameters of \textsc{OpenHands}.}
    \begin{adjustbox}{width=\linewidth,center}
	\begin{threeparttable}
		\begin{tabular}{lllll}
\toprule
\textbf{Configuration Option} & \textbf{Component} & \textbf{Type} & \textbf{Default} & \textbf{Range} \\ 
\midrule
\texttt{max\_message\_chars} & AI & Integer & 10000 & \{10000, 30000, 60000, 120000\} \\
\texttt{max\_input\_tokens} & AI & Integer & 16384 & \{16384, 32768, 65536, 131072, 200000\} \\
\texttt{max\_output\_tokens} & AI & Integer & 256 & \{256, 512, 1024, 2048, 4096\} \\
\texttt{temperature} & AI & Float & 0.0 & \{0.0, 0.2, 0.5, 0.8, 1.0\} \\
\texttt{top\_p} & AI & Float & 0.7 & \{0.7, 0.85, 0.95, 1.0\} \\
\texttt{disable\_vision} & AI & Boolean & \texttt{True} & \{\texttt{True}, \texttt{False}\} \\

\midrule
\texttt{caching\_prompt} & non-AI & Boolean & \texttt{True} & \{\texttt{True}, \texttt{False}\} \\
\texttt{timeout\_llm} & non-AI & Integer & 30 & \{30, 60, 120, 300\} \\
\texttt{num\_retries} & non-AI & Integer & 0 & \{0, 2, 5, 8\} \\
\texttt{retry\_min\_wait} & non-AI & Integer & 0 & \{0, 3, 5, 15\} \\
\texttt{retry\_max\_wait} & non-AI & Integer & 30 & \{30, 60, 120, 300\} \\
\texttt{retry\_multiplier} & non-AI & Float & 1.0 & \{1.0, 1.5, 2.0, 3.0\} \\
\texttt{native\_tool\_calling} & non-AI & Boolean & \texttt{True} & \{\texttt{True}, \texttt{False}\} \\
\texttt{function\_calling} & non-AI & Boolean & \texttt{False} & \{\texttt{False}, \texttt{True}\} \\
\texttt{enable\_browsing} & non-AI & Boolean & \texttt{False} & \{\texttt{False}, \texttt{True}\} \\
\texttt{enable\_history\_truncation} & non-AI & Boolean & \texttt{True} & \{\texttt{True}, \texttt{False}\} \\
\texttt{max\_iterations} & non-AI & Integer & 1 & \{1, 3, 5, 10\} \\
\texttt{max\_budget\_per\_task} & non-AI & Float & 0.0 & \{0.0, 0.005, 0.01, 0.05\} \\
\texttt{sandbox\_timeout} & non-AI & Integer & 60 & \{60, 120, 300\} \\
\texttt{use\_host\_network} & non-AI & Boolean & \texttt{False} & \{\texttt{False}, \texttt{True}\} \\
\texttt{enable\_auto\_lint} & non-AI & Boolean & \texttt{False} & \{\texttt{False}, \texttt{True}\} \\
\bottomrule
\end{tabular}
	\end{threeparttable}
    \end{adjustbox}
    \label{tab:openhands_knobs_table}
\end{table}


\begin{table}[h!]
	\centering
	\caption{Details of fidelity factors. $\boldsymbol{\mathcal{Z}}$ denotes the number of tested fidelity settings. The naming of each fidelity setting for the folders follow the same order as presented here, where the the value of each factor is separated by ``$-$''.}
    \label{tb:fidelity_space}
    \begin{adjustbox}{width=\linewidth,center}
	\begin{threeparttable}
		
\begin{tabular}{l|l|l|l}
\toprule
\textbf{LLM System}                          & \textbf{Fidelity   Factors} & \textbf{Values}      & \textbf{$\boldsymbol{\mathcal{Z}}$} \\ \midrule
 \multirow{4}{*}{\textbf{\textsc{LightRAG}}} 
 & \texttt{question-type}     & $\{bridge,comparison\}$        & \multirow{4}{*}{$216$} \\
 & \texttt{context-length}     & $\{400, 800, 1200, 1600\}$ &  \\
 & \texttt{supporting-facts-count} & $\{4,8,10\}$          &  \\
 & \texttt{context-count} & $\{2,3,4,5,6,7,8,9,10\}$          &  \\
 \hline
 \multirow{4}{*}{\textbf{\textsc{NaiveRAG}}} 
 & \texttt{question-ratio}      & $\{0.2, 0.5, 0.8\}$       & \multirow{4}{*}{$108$} \\
  & \texttt{corpus-scale}        & $\{0, 1, 2\}$             &  \\
 & \texttt{question-difficulty} & $\{easy, midian, hard\}$             &  \\
 & \texttt{dataset-category}    & $\{agriculture, art, biography, cs\}$          &  \\ 
\hline
\multirow{3}{*}{\textbf{\textsc{HtmlRAG}}} 
 & \texttt{data-count}     & $\{1, 2, 3, 4, 5, 6\}$        & \multirow{3}{*}{$90$} \\
 & \texttt{html-ratio}     & $\{0.1, 0.15, 0.2, 0.25, 0.3\}$ &  \\
 & \texttt{question-ratio} & $\{0.5, 0.75, 1.0\}$          &  \\
 \hline
 \multirow{5}{*}{\textbf{\textsc{vLLM}}} 
  & \texttt{request\_rate} & $\{1, 5, 10\}$ & \multirow{5}{*}{$216$}   \\
  & \texttt{burstiness} & $\{ 0.5, 1.0, 2.0\}$ &  \\
  & \texttt{max\_concurrency}     & $\{4, 8, 16, 32\}$ &  \\
  & \texttt{num\_prompts}     & $\{50, 100, 200\}$        \\
  & \texttt{repeat\_count} & $\{1, 2\}$          &  \\

 \hline
    \multirow{5}{*}{\textbf{\textsc{SGLang}}} 
 & \texttt{request\_rate}     & $\{5.0, 15.0, inf\}$       & \multirow{5}{*}{$108$} \\
 & \texttt{burstiness} & $\{ 1.0, 2.0\}$          &  \\
 & \texttt{max\_concurrency}     & $\{16, 32\}$       &  \\
 & \texttt{gsp\_num\_groups}     & $\{16, 32, 64\}$ &  \\
 & \texttt{gsp\_system\_prompt\_len} & $\{1024, 2048, 4096\}$ &  \\
 \hline

\multirow{4}{*}{\textbf{\textsc{AutoGPT}}} 
 & \texttt{task\_type}     
 & $\{simple, moderate, complex, multi\_stage\}$        
 & \multirow{3}{*}{$112$} \\
 & \texttt{requests\_count}     
 & $\{1, 2, 3, 4\}$ 
 &  \\
 & \texttt{workload\_category} 
 & \makecell[l]{
     $\{math\_reasoning, code\_generation, logic\_puzzles, data\_analysis,$\\
     $memory\_retrieval, instruction\_adherence, text\_classification\}$
   } 
 &  \\ \hline

 \multirow{4}{*}{\textbf{\textsc{OpenHands}}} 
 & \texttt{facts-count}   & $\{7,8,9,10,...,20\}$ & \multirow{4}{*}{$82$} \\
 & \texttt{rules-count}   & $\{1,2,3,4,5,6,7,8,9\}$ &  \\
 & \texttt{proof-depth}   & $\{1,2,5\}$ &  \\
 & \texttt{support-count} & $\{1,3,6\}$ &  \\

\bottomrule
\end{tabular}

	\end{threeparttable}
    \end{adjustbox}
\end{table}


\begin{table}[h!]
	\centering
	\caption{Details of the inference objectives.}
    \label{tb:obj}
    \begin{adjustbox}{width=\linewidth,center}
	\begin{threeparttable}
		
\begin{tabular}{l|l|l}
\toprule
\textbf{LLM System}                           & \textbf{Objectives} & \textbf{Description}    \\ \midrule
\multirow{9}{*}{\textbf{LightRAG}} 
 & \texttt{mrr} & Mean reciprocal rank. \\
 & \texttt{ndcg} & Ranking quality of retrieved contexts. \\
 & \texttt{context\_similarity} & Semantic similarity between query and retrieved contexts. \\
 & \texttt{f1} & Token-level F1 score measuring overlap between prediction and ground truth. \\
& \texttt{best\_match\_position} & Rank position of the first relevant document in the retrieved list. \\
 & \texttt{relevant\_docs\_count} & Number of retrieved documents exceeding the relevance similarity threshold. \\
 & \texttt{precision} & Token-level precision: proportion of predicted tokens that are correct. \\
 & \texttt{recall} & Token-level recall: proportion of ground truth tokens that are recovered. \\
 & \texttt{test\_time} & Time spent evaluating test questions \\
 \hline
\multirow{8}{*}{\textbf{NaiveRAG}} 
 & \texttt{mrr} & Mean reciprocal rank. \\
 & \texttt{ndcg} & Ranking quality of retrieved contexts. \\
 & \texttt{context\_similarity} & Semantic similarity between query and retrieved contexts. \\
 & \texttt{lexical\_ac} & Lexical answer correctness of generated responses. \\
 & \texttt{answer\_precision} & Precision of generated answers. \\
 & \texttt{answer\_llmaaj} & LLM-as-a-judge correctness score. \\
 & \texttt{avg\_similarity} & Average semantic similarity score. \\
  & \texttt{test\_time} & Time spent evaluating test questions. \\
 \hline
 \multirow{8}{*}{\textbf{HtmlRAG}} 
 & \texttt{mrr} & Mean reciprocal rank. \\
 & \texttt{ndcg} & Ranking quality of retrieved contexts. \\
 & \texttt{context\_similarity} & Semantic similarity between query and retrieved contexts. \\
 & \texttt{lexical\_ac} & Lexical answer correctness of generated responses. \\
 & \texttt{answer\_precision} & Precision of generated answers. \\
 & \texttt{answer\_llmaaj} & LLM-as-a-judge correctness score. \\
 & \texttt{avg\_similarity} & Average semantic similarity score. \\
  & \texttt{test\_time} & Time spent evaluating test questions. \\
\hline

\multirow{5}{*}{\textbf{vLLM}} 
 & \texttt{throughput} & Tokens processed per second. \\
 & \texttt{ttft} & Time to first token. \\
 & \texttt{tpot} & Token processing time per generated token. \\
 & \texttt{normalized\_latency} & End-to-end latency normalized by output length. \\
 & \texttt{success\_ratio} & Fraction of successful requests. \\
\hline
\multirow{5}{*}{\textbf{SGLang}} 
 & \texttt{throughput} & Tokens processed per second. \\
 & \texttt{ttft} & Time to first token. \\
 & \texttt{tpot} & Token processing time per generated token. \\
 & \texttt{normalized\_latency} & End-to-end latency normalized by output length. \\
 & \texttt{success\_ratio} & Fraction of successful requests. \\
\hline

\multirow{5}{*}{\textbf{AutoGPT}} 
 & \texttt{success\_rate} & Fraction of tasks completed successfully. \\
 & \texttt{timeout\_rate} & Aggregate timeout fraction. \\
 & \texttt{step\_timeout\_rate} & Fraction of tasks with per-step timeout. \\
 & \texttt{avg\_steps\_per\_wall\_timeout} & Average steps before a wall timeout. \\
 & \texttt{instruction\_adherence} & Degree of constraint and directive compliance. \\ 
 \hline

 \multirow{5}{*}{\textbf{OpenHands}} 
 & \texttt{accuracy} & final answer matches the gold answer. \\
 & \texttt{evidence\_recall} & \makecell[l]{Gold evidence participation rate measures the fraction of gold supporting \\evidence used in the model's reasoning.} \\
 & \texttt{proof\_step\_deviation:} & \makecell[l]{Proof-step difference measures the absolute gap between the generated \\and gold proof lengths.} \\

 \bottomrule
\end{tabular}

    \end{threeparttable}
    \end{adjustbox}
\end{table}


\begin{table}[h!]
	\centering
	\caption{Details of the cost of hyperparameter configuration measurements.}
    \label{tb:cost}
    \begin{adjustbox}{width=\linewidth,center}
    \begin{threeparttable}
        
\begin{tabular}{l|l|l}
\toprule
\textbf{LLM System}                           & \textbf{Cost} & \textbf{Description}    \\ \midrule

\multirow{10}{*}{\textbf{LightRAG}} 
& \texttt{total\_time} & Total pipeline execution time. \\
& \texttt{insert\_time} & Time spent on document insertion (indexing phase). \\
& \texttt{insert\_input} & Number of input tokens consumed during insertion. \\
& \texttt{insert\_output} & Number of output tokens generated during insertion. \\
& \texttt{insert\_total} & Total tokens (input + output) used in insertion. \\
& \texttt{insert\_calls} & Number of API calls made during insertion. \\
& \texttt{query\_input} & Number of input tokens consumed during query processing. \\
& \texttt{query\_output} & Number of output tokens generated during query processing. \\
& \texttt{query\_total} & Total tokens (input + output) used in query processing. \\
& \texttt{query\_calls} & Number of API calls made during query processing. \\
\hline
 
\multirow{7}{*}{\textbf{NaiveRAG}} 
& \texttt{total\_time} & Total pipeline execution time. \\
& \texttt{build\_time} & Time spent building the retrieval index or database. \\
& \texttt{chunk\_time} & Time spent chunking input documents. \\
& \texttt{total\_tokens} & Total number of consumed tokens. \\
& \texttt{embedding\_tokens} & Number of tokens consumed by the embedding model. \\
& \texttt{llm\_tokens} & Number of tokens consumed by the generation model. \\
& \texttt{avg\_tokens\_per\_question} & Average token usage per question. \\ \hline
\multirow{7}{*}{\textbf{HtmlRAG}} 
& \texttt{total\_time} & Total pipeline execution time. \\
& \texttt{build\_time} & Time spent building the retrieval index or database. \\
& \texttt{chunk\_time} & Time spent chunking input documents. \\
& \texttt{total\_tokens} & Total number of consumed tokens. \\
& \texttt{embedding\_tokens} & Number of tokens consumed by the embedding model. \\
& \texttt{llm\_tokens} & Number of tokens consumed by the generation model. \\
& \texttt{avg\_tokens\_per\_question} & Average token usage per question. \\
\hline

\multirow{2}{*}{\textbf{vLLM}} 
& \texttt{process\_cpu\_seconds\_stats} & Total CPU time consumed by the vLLM process. \\
& \texttt{benchmark\_duration\_s} & End-to-end benchmark duration. \\
\hline
\multirow{2}{*}{\textbf{SGLang}} 
& \texttt{process\_cpu\_seconds\_stats} & Total CPU time consumed by the SGLang process. \\
& \texttt{benchmark\_duration\_s} & End-to-end benchmark duration. \\
\hline
\multirow{3}{*}{\textbf{AutoGPT}} 
& \texttt{prompt\_tokens} & Input tokens consumed. \\
& \texttt{completion\_tokens} & Output tokens consumed. \\
& \texttt{total\_duration\_s} & Total wall-clock time for all tasks. \\
\hline

\multirow{6}{*}{\textbf{OpenHands}} 
& \texttt{total\_time} & Total wall-clock time consumed by the task. \\
& \texttt{cput\_time} & Total CPU time consumed during task execution. \\
& \texttt{total\_tokens} & Total number of tokens consumed, including input and output tokens. \\
& \texttt{input\_tokens} & Number of tokens used as model input. \\
& \texttt{output\_tokens} & Number of tokens generated by the model. \\
& \texttt{peak\_memory\_usage} & Peak memory usage during task execution. \\

 \bottomrule

\end{tabular}

    \end{threeparttable}
    \end{adjustbox}
\end{table}


\begin{table}[h!]
	\centering
	\caption{Details of the hardware metrics used for all LLM systems.}
    \label{tb:hardware}
		\begin{tabular}{l|l}
\toprule \textbf{Hardware} & \textbf{Description}    \\ \midrule
\texttt{CPU Utilization}               & Percentage of CPU resources currently in use.              \\
\texttt{RAM Utilization}               & Percentage of system memory currently occupied.        \\ 
\texttt{GPU Core Utilization}          & Percentage of GPU compute cores actively in use.        \\
\texttt{VRAM Utilization}              & Percentage of GPU memory currently occupied.        \\
\texttt{GPU Temperature}               & Current temperature of the GPU in degrees Celsius.        \\
\texttt{GPU Power Consumption}         & Current power usage of the GPU in watts.        \\
\bottomrule
\end{tabular}

\end{table}

\clearpage

\lstdefinestyle{logtrace}{
    basicstyle=\ttfamily\scriptsize\linespread{0.92}\selectfont,
    backgroundcolor=\color{codebg},
    showstringspaces=false,
    frame=single,
    framerule=0.3pt,
    breaklines=true,
    breakatwhitespace=false,
    columns=fullflexible,
    xleftmargin=4pt,
    xrightmargin=4pt,
    aboveskip=4pt,
    belowskip=6pt
}

\begin{figure}[H]
\centering
\begin{minipage}{0.98\linewidth}
\begin{lstlisting}[style=logtrace]
================================================================
2026-04-30T00:00:30+08:00  AutoGPT Agent Protocol launch
  cwd      : <workspace>/AutoGPT/classic/original_autogpt
  command  : poetry run serve
  log path : <benchmark-log-dir>/autogpt_server.log
================================================================

--- AutoGPT server ---
<python-env>/site-packages/pydantic/_internal/_fields.py:160:
  UserWarning: Field "model_base_url_map" has conflict with protected namespace "model_".
2026-04-30 00:00:32,264 INFO     HTTP Request: GET http://localhost:8080/v1/models "HTTP/1.1 200 OK"
2026-04-30 00:00:32,409 WARNING  Frontend not found. <workspace>/AutoGPT/classic/frontend/build/web does not exist. The frontend will not be available.
2026-04-30 00:00:32,409 INFO     AutoGPT server starting on http://localhost:8000
2026-04-30 00:00:56,436 INFO     Configure google_api_key and custom_search_engine_id to use Google API search.
2026-04-30 00:00:57,828 INFO     HTTP Request: POST http://localhost:8080/v1/chat/completions "HTTP/1.1 200 OK"
2026-04-30 00:00:57,831 WARNING  Parsing attempt #1 failed: InvalidAgentResponseError: Assistant did not use a tool
2026-04-30 00:00:59,384 INFO     HTTP Request: POST http://localhost:8080/v1/chat/completions "HTTP/1.1 200 OK"
2026-04-30 00:00:59,386 WARNING  Parsing attempt #2 failed: InvalidAgentResponseError: Assistant did not use a tool
...
2026-04-30 00:01:00,814 INFO     HTTP Request: POST http://localhost:8080/v1/chat/completions "HTTP/1.1 200 OK"
2026-04-30 00:01:00,817 WARNING  Parsing attempt #3 failed: InvalidAgentResponseError: Assistant did not use a tool
2026-04-30 00:01:02,145 INFO     HTTP Request: POST http://localhost:8080/v1/chat/completions "HTTP/1.1 200 OK"

--- vLLM API server ---
(APIServer pid=<pid>) INFO:     Started server process [<pid>]
(APIServer pid=<pid>) INFO:     Application startup complete.
(APIServer pid=<pid>) INFO:     127.0.0.1:<port> - "GET /v1/models HTTP/1.1" 200 OK
(APIServer pid=<pid>) INFO 04-30 00:00:57 [loggers.py:259] Engine 000: Avg prompt throughput: 38.3 tokens/s, Avg generation throughput: 11.9 tokens/s, Running: 1 reqs, Waiting: 0 reqs, GPU KV cache usage: 0.3%, Prefix cache hit rate: 0.0%
(APIServer pid=<pid>) INFO:     127.0.0.1:<port> - "POST /v1/chat/completions HTTP/1.1" 200 OK
...
(APIServer pid=<pid>) INFO 04-30 00:01:07 [loggers.py:259] Engine 000: Avg prompt throughput: 20.0 tokens/s, Avg generation throughput: 96.0 tokens/s, Running: 1 reqs, Waiting: 0 reqs, GPU KV cache usage: 0.3%, Prefix cache hit rate: 82.6%
\end{lstlisting}
\end{minipage}
\caption{Sanitized logging trace of the LLM system \textsc{AutoGPT}.}
\label{fig:autogpt-log}
\end{figure}





\end{document}